%% file: rahul_etal_RAL_19.tex
\documentclass[letterpaper, 10 pt, journal, twoside]{IEEEtran}  

\IEEEoverridecommandlockouts                              

\usepackage{enumitem}
\usepackage{graphics} 
\usepackage{epsfig} 
\usepackage{mathptmx} 
\usepackage{times} 
\usepackage{amsmath} 
\usepackage{amssymb}  
\usepackage{graphicx}
\usepackage{tabularx}
\usepackage{multicol}
\usepackage{algorithm}
\usepackage{algorithmic}
\usepackage{multirow}
\usepackage{todonotes}%
\usepackage{lettrine}
\usepackage{subcaption}
\usepackage{wasysym}
\usepackage{bm}

\usepackage[font={scriptsize,it}]{caption} 
\usepackage[linewidth=1pt]{mdframed}
\usepackage{setspace}

\newcommand{\argmin}[1]{\underset{#1}{\operatorname{\mathbf{arg}}\,\operatorname{\mathbf{min}}}\;}
\newcommand{\x}{\mathbf{x}}
\newcommand{\xproj}{\mathbf{\check{x}}}

\newcommand{\ao}{\mathbf{u}}
\newcommand{\X}{\mathbf{X}}

\newcommand{\varthet}{\sigma_{\theta}^2}
\newcommand{\varph}{\sigma_{\phi}^2}
\newcommand{\varr}{\sigma_{r}^2}

\graphicspath{{pictures/}}
\usepackage{cite}
\usepackage[bookmarks=false]{hyperref}
\usepackage{leftidx}

\definecolor{lightlightgrey}{rgb}{0.9,0.9,0.9}
\definecolor{Red}{rgb}{1,0,0}
\definecolor{Blue}{rgb}{0,0,1}
\definecolor{Green}{rgb}{0,1,0}
\definecolor{magenta}{rgb}{1,0,.6}
\definecolor{lightblue}{rgb}{0,.5,1}
\definecolor{lightpurple}{rgb}{.6,.4,1}
\definecolor{gold}{rgb}{.6,.5,0}
\definecolor{orange}{rgb}{1,0.4,0}
\definecolor{hotpink}{rgb}{1,0,0.5}
\definecolor{newcolor2}{rgb}{.5,.3,.5}
\definecolor{newcolor}{rgb}{0,.3,1}
\definecolor{newcolor3}{rgb}{1,1,1}
\definecolor{darkgreen1}{rgb}{0, .35, 0}
\definecolor{darkgreen}{rgb}{0, .6, 0}
\definecolor{darkred}{rgb}{.75,0,0}

\xdefinecolor{olive}{cmyk}{0.64,0,0.95,0.4}
\xdefinecolor{purpleish}{cmyk}{0.75,0.75,0,0}

\xdefinecolor{red}{rgb}{1,0,0}

\usepackage{float}
\floatstyle{boxed}
\newfloat{math}{thp}{lop}
\floatname{math}{Equation}

\begin{document}

\title{Active Perception based Formation Control for Multiple Aerial Vehicles}

\author{Rahul Tallamraju$^{1,3}$, Eric Price$^1$, Roman Ludwig$^1$, Kamalakar Karlapalem$^3$, \\ Heinrich H.\ B\"ulthoff$^2$,  Michael J.\ Black$^1$ and Aamir Ahmad$^1$%
\thanks{$^{1}$Max Planck Institute for Intelligent Systems, T\"ubingen, Germany.
        {\tt\footnotesize {rahul.tallamraju, eprice, roman.ludwig, black, aamir.ahmad}@tuebingen.mpg.de}}%
\thanks{$^{2}$Max Planck Institute for Biological Cybernetics, T\"ubingen, Germany.
        {\tt\footnotesize hhb@tuebingen.mpg.de}}%
\thanks{$^{3}$Agents and Applied Robotics Group, IIIT Hyderabad, India.
        {\tt\footnotesize kamal@iiit.ac.in}}%
\thanks{Manuscript Accepted to IEEE Robotics and Automation Letters 2019. }
\thanks{DOI: 10.1109/LRA.2019.2932570}
\thanks{The arxiv manuscript has an additional figure, Figure 5, which has been cited in the published RA-L manuscript.}
}


\maketitle

\begin{abstract}
\input{abstract}
\end{abstract}


\IEEEpeerreviewmaketitle

\input{intro_new}

\input{related_work}

\input{poa}

\input{experiments}

\input{sim_experiments}

\input{conclusions}

\section*{Acknowledgments}
\noindent The authors would like to thank Igor Martinovic and Mason Landry for their help with the field experiments.
For the published RA-L manuscript and hence this revision on Arxiv, we thank the anonymous reviewers, associate editor and the editor Eric Marchand for their highly constructive feedback.	

\bibliographystyle{IEEEtran}
\bibliography{paper}

\end{document}

%% file: abstract.tex
We present a novel robotic front-end for autonomous aerial motion-capture (mocap) in outdoor environments. In previous work, we presented an approach for cooperative detection and tracking (CDT) of a subject using multiple micro-aerial vehicles (MAVs). However, it did not ensure optimal view-point configurations of the MAVs to minimize the uncertainty in the person's cooperatively tracked 3D position estimate. 
In this article, we introduce an active approach for CDT. In contrast to cooperatively tracking only the 3D positions of the person, the MAVs can actively compute optimal local motion plans, resulting in optimal view-point configurations, which minimize the uncertainty in the tracked estimate. We achieve this by decoupling the goal of active tracking into a quadratic objective and non-convex constraints corresponding to angular configurations of the MAVs w.r.t.\ the person. We derive this decoupling using Gaussian observation model assumptions within the CDT algorithm. We preserve convexity in optimization by embedding all the non-convex constraints, including those for dynamic obstacle avoidance, as external control inputs in the MPC dynamics. Multiple real robot experiments and comparisons involving 3 MAVs in several challenging scenarios are presented. 

%% file: intro_new.tex
\section{Introduction}



\IEEEPARstart{A}{erial} motion capture (mocap) of humans in unstructured outdoor scenarios is a challenging and important problem \cite{eprice,huang2018act,zhou2018human,xu2018flycap,nageli2018flycon}. It may directly facilitate applications like sports medicine, cinematography, or search and rescue operations. The front-end of our mocap system consists of a team of micro aerial vehicles (MAVs), autonomously detecting, tracking and following a person. The front-end is responsible for the online task of the system, which is to continuously estimate the 3D global position of the person and keep him/her centered in the field of view of their on-board camera, while he/she performs activities such as walking, running, jumping, etc. The online task is the core focus of this article. The offline task, which is not addressed in this article, is to estimate full body skeleton pose and shape using images acquired by the MAVs. 

\begin{figure} [t]
	\centering
	\includegraphics[width=\columnwidth]{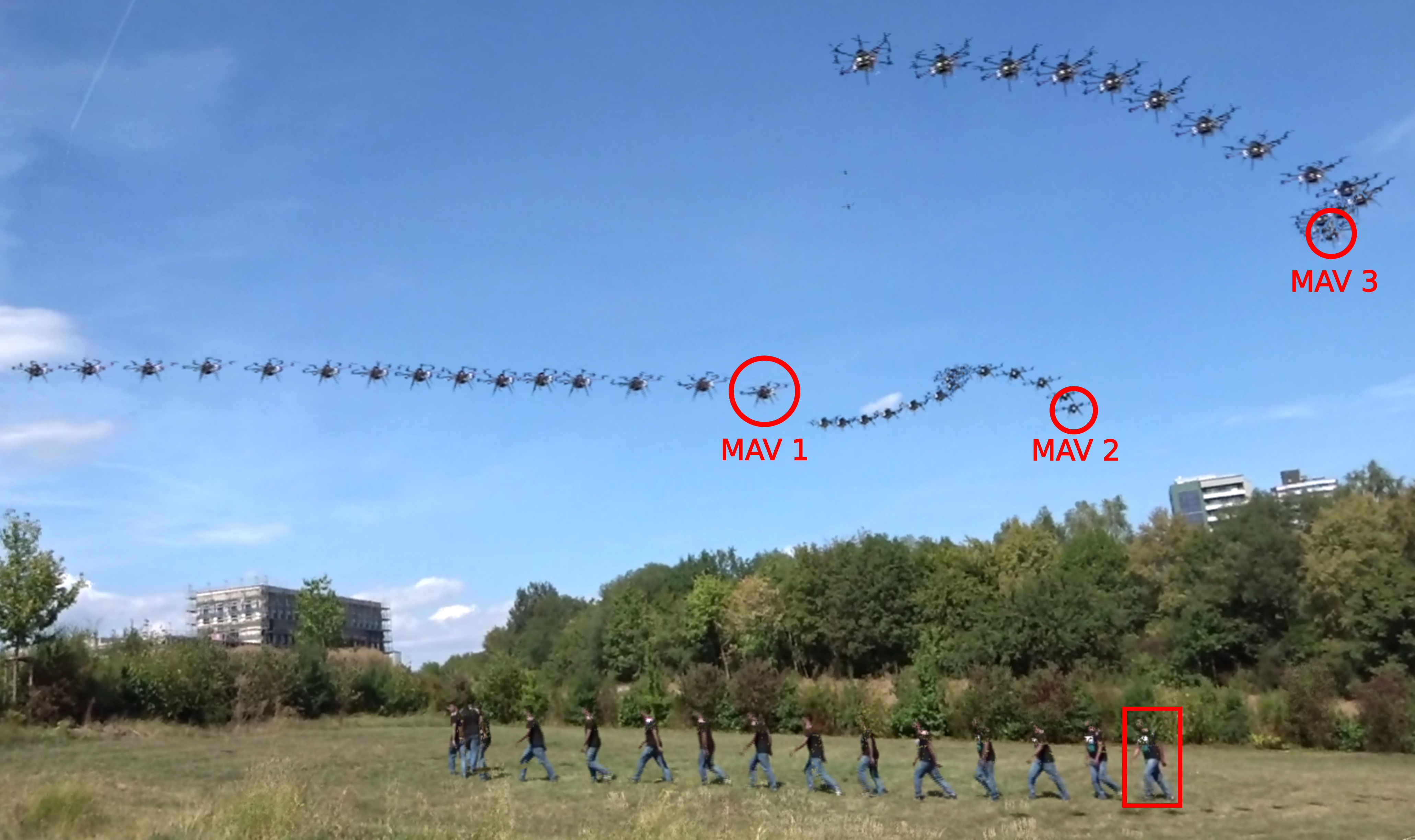}
	\caption{Multi-exposure image of three octocopter MAVs cooperatively tracking a fast walking person while maintaining a perception-driven formation using the approach proposed in this paper. The person walks from right to left. The initial positions of the MAVs in their trajectories are circled.} 
	\label{cover}
\end{figure} 


In order to develop the mocap front-end, in our previous work \cite{eprice}, we presented a marker-less deep neural network (DNN) based cooperative detection and tracking (CDT) algorithm. That work had two major shortcomings. First, the motion planner on each MAV did not facilitate uncertainty minimization of the person's 3D position estimate. Second, the local motion plans generated by the MAVs using a model-predictive controller (MPC) were subsequently modified using potential fields for collision avoidance. This further led to sub-optimal trajectories and robots getting easily stuck in field local minima, e.g., stuck behind each other or behind closely spaced obstacles.



In the current work, we solve both of the aforementioned problems through a novel convex decentralized formation controller based on MPC.
This new MPC actively minimizes the joint uncertainty in the tracked person's 3D position estimate while following him/her. The key novelties are as follows.
\begin{itemize}[leftmargin=*]
\item The first novel idea, which enables us to formulate this problem as a locally convex MPC, is the decoupling of the joint uncertainty minimization into i) a convex quadratic objective that maintains a threshold distance to the tracked person, and ii) constraints that enforce angular configurations of the MAVs with respect to (w.r.t.) the person. 
We derive this decoupling based on Gaussian observation model assumptions used within the CDT algorithm. 
\
\item To guarantee the safety of the motion plans, we incorporate collision avoidance constraints w.r.t.\ i) other MAVs, ii) the tracked person and iii) static obstacles, only as locally convex constraints. Collision avoidance and angular configuration constraints are inherently non-convex. We preserve convexity in our MPC formulation by converting them to external control input terms embedded inside the MPC dynamics, which are explicitly computed at every iteration of the MPC. 
\end{itemize}


\begin{figure} [t]
	\centering
	\includegraphics[width=\columnwidth]{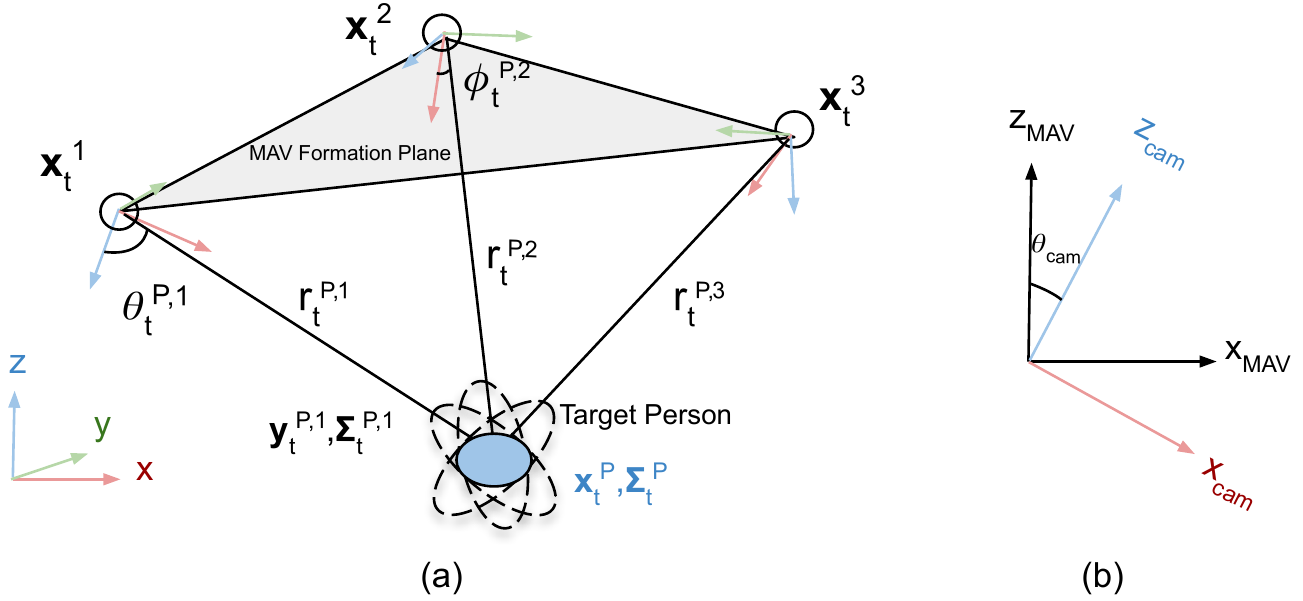}
	\caption{Formation of $3$ MAVs with the mathematical notations used in this paper.}
	\label{fig:spher2cart}
\end{figure}


Real robot experiments and comparisons involving 3 MAVs in several challenging scenarios are presented. Figure~\ref{cover} shows a multi-exposure image of a short sequence from one of these experiments. Simulation results with up to 16 robots demonstrate the robustness of our approach and scalability w.r.t.\ to the number of robots. We provide open-source ROS-based\footnote{\url{https://github.com/AIRCAP/AIRCAP}} source code of our method.

%% file: related_work.tex
\section{State-of-the-Art}
\label{sec:sota}

\textit{Multi-robot active tracking:} 
In \cite{morbidi2011active}, multi-robot trajectories are analytically derived and uncertainty bounds are identified for optimizing the fused estimate of the target position. 
In \cite{zhou2011multirobot}, a non-convex optimization problem is solved analytically, to minimize the trace of the fused target covariance. However, their approach is centralized and uses Gauss-Seidel relaxation to compute optimal solutions for multiple robots. 
Centralized non-convex optimization \cite{hausman2015cooperative, hausman2016occlusion} is used to track a target in stochastic environments and locally minimize the fused estimate of the target position. 
In \cite{lima2015formation}, active perception based formation control is addressed  using a decentralized non-linear MPC. However, their method only identifies sub-optimal control inputs due to the non-convexity of optimization and the lack of collision avoidance guarantees. 
In \cite{falanga2018pampc}, a perception-aware MPC generates real-time motion plans which maximize the visibility of a desired target. The motion plans are, however, generated only for a single aerial robot. 
A marker-based multi-robot aerial motion capture system is presented in \cite{nageli2018flycon}, where the formation control is achieved using a decentralized non-linear MPC. Scalability in the number of robots and formation collision avoidance behaviors are not explicitly addressed in their approach.

\textit{Obstacle Avoidance}: Sequential convex programming is applied to solve centralized multi-robot collision-free trajectory generation \cite{augugliaro2012generation}. Due to non-convexities which arise from obstacle avoidance constraints, the obtained solutions are locally optimal albeit fast for a given time horizon. In \cite{lima2015formation} repulsive potential field functions are employed as one of the optimization objectives to avoid collisions. This is highly limiting as obstacle avoidance cannot be guaranteed.  Recent work in distributed multi-agent obstacle avoidance \cite{alonso2015collision} convexifies the reciprocal velocity obstacle constraint, to characterize velocities that do not lead to a collision in a perfectly localized environment. Motivated by multi-view cinematography applications, distributed non-linear model predictive control \cite{nageli2017} is used to identify locally optimal motion plans for aerial vehicles. In one of our previous works \cite{rahult}, we developed a convex optimization program to generate local collision-free motion plans, while tracking a movable pick and place static target using multiple aerial vehicles. This approach generates fast, feasible motion plans and has a linear computational complexity $(O(n))$ in the number of environmental obstacles. 
The work was validated only in a simulation environment and, moreover, obstacle avoidance was not guaranteed. In our current work, we consider real robots, stochasticity in the environment and bounds on repulsive potential fields to guarantee collision avoidance and generate safe local motion plans.

%% file: poa.tex
\section{Proposed Approach}
\label{sec:poa}

\subsection{Preliminaries and Problem Statement} \label{prelim}

Let there be K MAVs tracking a person P. 
The pose (position and orientation) of the $k^{th}$ MAV in the  world frame
at time $t$ is given by $\xi_t^{k} = [(\mathbf{x}_t^{k})^\top ~ (\Theta_t^{k})^\top] =  [ (\ \mathbf{x}_t^{k})^\top ~ [\psi_t^{k} ~ \theta^k_t ~ \phi_t^k]^\top] \in \mathbb{R}^6$. The MAVs pose uncertainty covariance matrix is denoted as  $\boldsymbol{\Sigma}_t^{k} \in \mathbb{R}^{6\times6}$. Each MAV has an on-board, monocular, perspective camera. It is important to note that the camera is rigidly attached to the MAV's body frame, pitched down at an angle of $\theta_\textrm{cam} = 45^\circ$. Hence, the pose of the camera in the MAV's frame is $[0 ~ 0 ~ 0 ~  0 ~ \theta_\textrm{cam} ~ 0]^\top$ (see Fig.~\ref{fig:spher2cart}).
The observation measurement of the person $P$'s position made by MAV $k$ in its camera frame is given by a mean range $r_t^{P,k}$, bearing $\phi_t^{P,k}$ and inclination $\theta^{P,k}_t$ (see Fig. \ref{fig:spher2cart}) in spherical coordinates. These are associated with a zero mean measurement noise, denoted by $\sigma^{P,k}_{r,t}$, $\sigma^{P,k}_{\phi,t}$ and $\sigma^{P,k}_{\theta,t}$ respectively. We assume that these measurements are uncorrelated.
In camera frame's Cartesian coordinates these measurements are denoted as $\mathbf{y}_t^{P,k} \in\mathbb{R}^3$ and $\boldsymbol{\Sigma}_t^{P,k} \in \mathbb{R}^{3\times3}$ respectively. 
The fused estimate of the person $P$'s position and uncertainty covariance in the world frame are denoted as $\mathbf{x}_t^P$  and $\boldsymbol{\Sigma}_t^{P}$. These estimates are computed by each MAV $k$ fusing its own measurements, $\mathbf{y}_t^{P,k}$, with the measurements received from all other teammates ($\mathbf{y}_t^{P,j} \forall j=[1...K]; j\neq k $  ).   

The MAVs operate in an  environment with $M$ known static obstacles (our approach is agnostic to how these obstacles are detected) and $K-1$ neighboring MAVs as dynamic obstacles. The goal of each MAV is to cooperatively track the person using a replicated instance of the proposed formation control algorithm. This involves (a) minimizing the MAV's fused estimate ${\boldsymbol{\Sigma}}_t^{P}$ of the measurement uncertainty covariance and, (b) avoiding $M+K-1$ environmental obstacles.

\begin{figure} [t]
	\centering
	\includegraphics[width=\columnwidth]{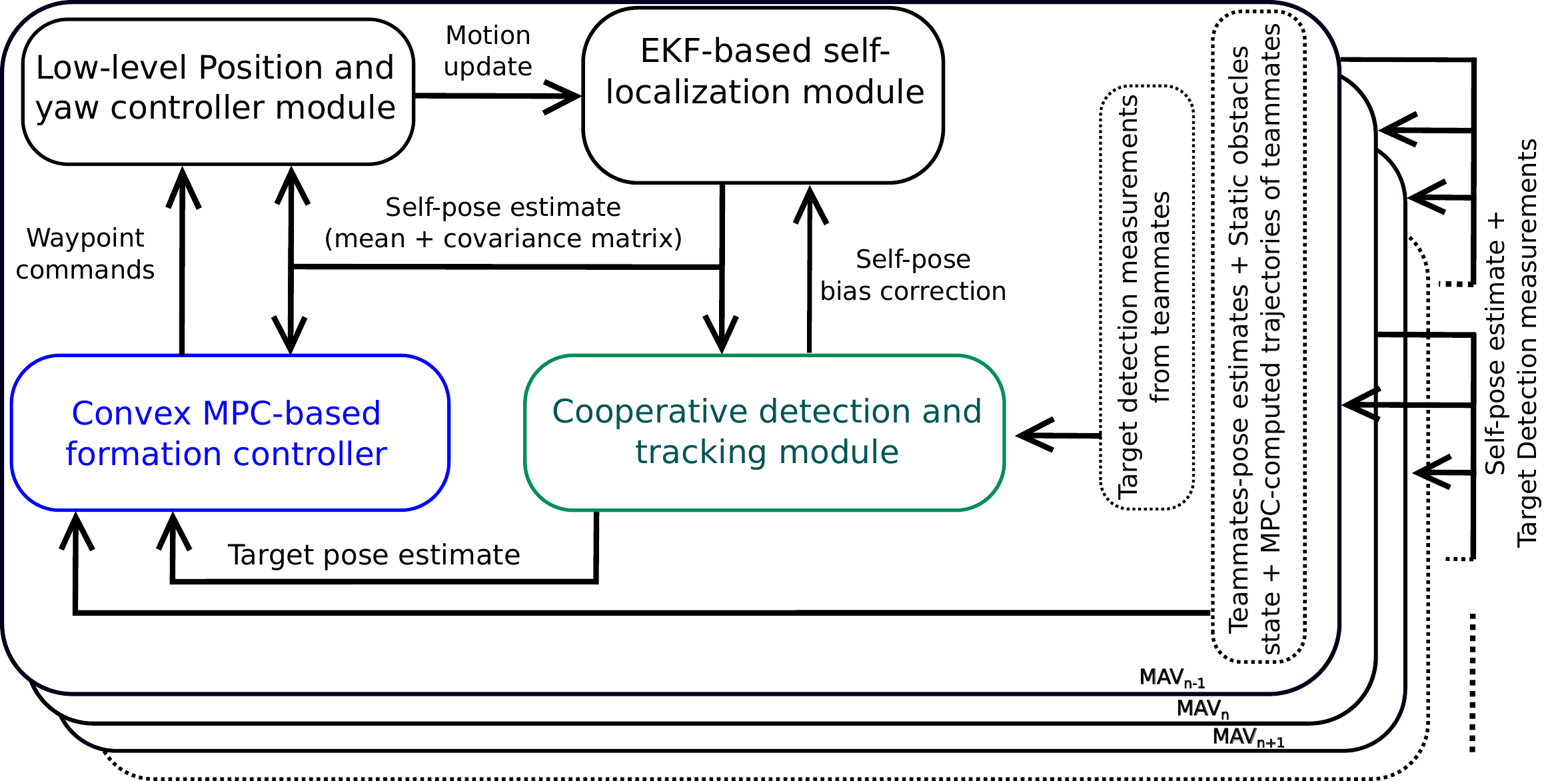}
	\caption{Our aerial mocap front-end architecture. This paper focuses on the blue module.}
	\label{CDT_fig}
\end{figure}

\subsection{Cooperative Detection and Tracking} \label{CDT}

Fig.~\ref{CDT_fig} describes our mocap front-end architecture including i) the decentralized formation controller and ii) cooperative detection and tracking (CDT) modules.
In \cite{eprice}, the focus was on developing the CDT algorithm (green block in Fig.~\ref{CDT_fig}). The work in the current paper focuses on developing the formation controller (blue block in Fig.~\ref{CDT_fig}), which utilizes information from CDT and other modules.
In CDT, each MAV runs its own instance of an EKF for fusing detections $ \in \mathbb{R}^3$ made by it and all its teammates. This results in a replicated state estimate of the person at each MAV, which is used to predict a region of interest (ROI) in future camera frames. The ROI guides the yaw of each MAV thereby ensuring that the tracked person is in the MAV's field of view. In Sec.~\ref{DQMPC_sec}, we use these state estimates to drive our decentralized formation controller.  At every time step, the controller of MAV $k$ generates way-point positions and velocities for itself using (a) estimated state of the tracked person, (b) horizon motion plans communicated by teammates and (c) the positions of obstacles. The way-points guide MAV $k$'s low-level position and yaw controller.

\begin{figure}[!htp]
	\centering
	\includegraphics[width=\columnwidth]{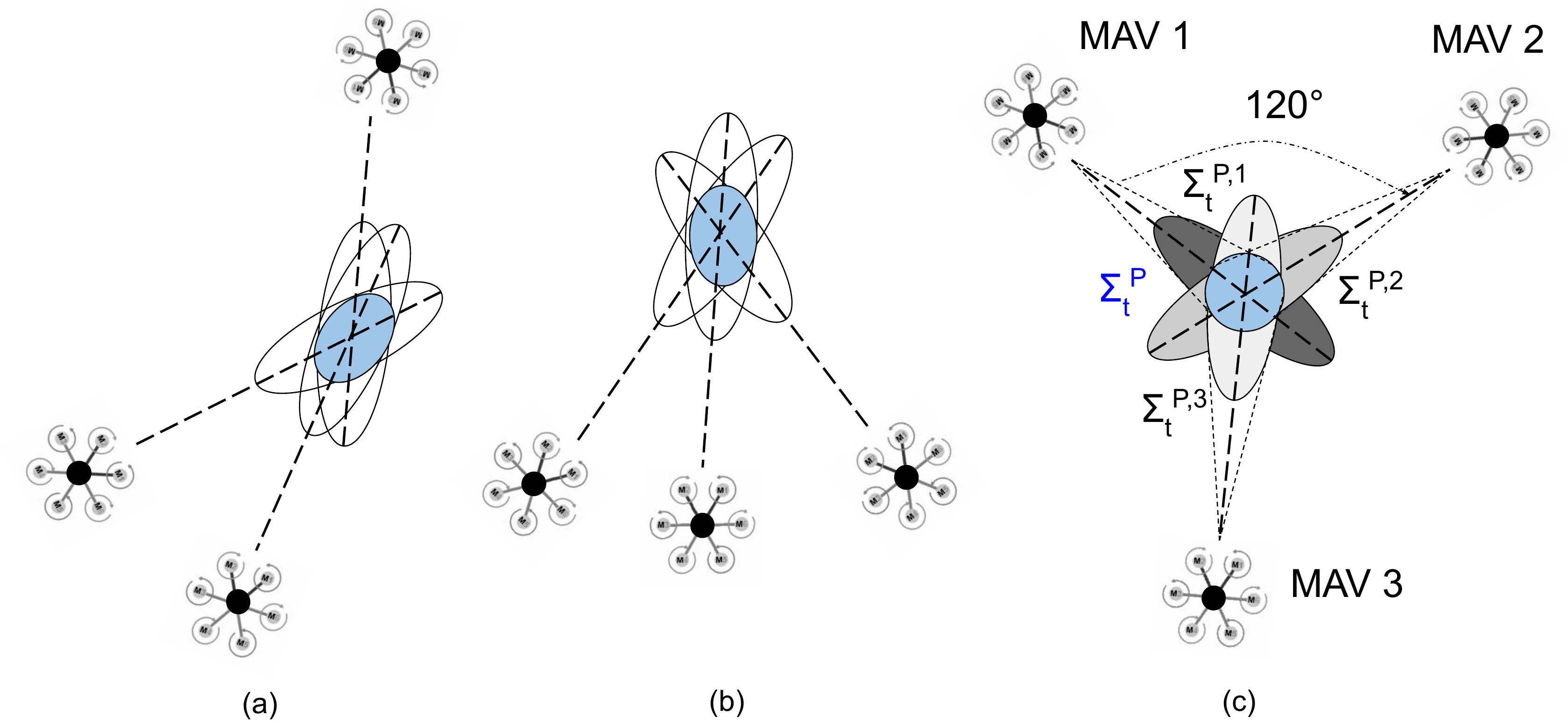}   
	\caption{(a), (b) represent arbitrary configurations with large joint uncertainties. (c) is the optimal configuration for 3 MAVs used in this work.}     
	\label{f:act_angle}
\end{figure}

\subsection{Measurement Models and Joint Uncertainty Minimization} \label{acttrack}

\subsubsection{Measurement Model for a Single MAV}

For a MAV $k$ in this sub-section we drop the superscripts and subscripts.
In a supplementary multimedia file attached with this paper, we empirically show that (a) the variance of the noise in the range measurement evolves quadratically with distance to the person, i.e., $\sigma_{r}^2 = C_1 r^2$, and (b) the variance of the noise in both bearing $\phi$ and inclination $\theta$ measurements are approximately
constant w.r.t.\ the distance to the person, which we represent as $\sigma_{\phi}^2 = C_2$ and $\sigma_{\theta}^2 = C_3$. $C_1, C_2$ and $C_3$ are positive constants specific to the system.

We now convert the measurement noise variances from spherical to Cartesian coordinates.
When using converted measurements (especially when measurement noise covariance
is explicitly required) and performing estimation in Cartesian coordinates, most
previous works, e.g., \cite{zhao_best_linear}, \cite{Rong_Li_survey_measurments_part_III}, consider the measurement-conditioned conversion
approach as derived in \cite{l_mo_bar_shalom_unbiased_converted} or \cite{miller_coordinate_transform} as one of the best approximations. 
Using this conversion from \cite{l_mo_bar_shalom_unbiased_converted} or \cite{miller_coordinate_transform} the noise covariances in Cartesian coordinates, i.e., the elements of $\boldsymbol{\Sigma}_t^{P,k}$ can be written as 

\begin{footnotesize}
\begin{eqnarray}
\label{eqn:single_mav_model}
\sigma_x^2  &=& A_1f_1(\varthet,\varph)g_1(\theta,\phi)r^2 + B_1h_1(\theta,\phi,\varthet,\varph)(r^2 + \varr), \nonumber\\  
\sigma_y^2  &=& A_2f_2(\varthet,\varph)g_2(\theta,\phi)r^2 + B_2h_2(\theta,\phi,\varthet,\varph)(r^2 + \varr) \nonumber\\  
\sigma_z^2  &=& A_3f_3(\varthet)r^2 + B_3h_3(\theta,\varthet)(r^2 + \varr) \nonumber\\  
\sigma_{xy}^2  &=& A_4f_4(\varthet,\varph)g_4(\theta,\phi)r^2 + B_4h_4(\theta,\phi,\varthet,\varph)(r^2 + \varr), \nonumber\\  
\sigma_{xz}^2  &=& A_5f_5(\varthet,\varph)g_5(\theta,\phi)r^2 + B_5h_5(\theta,\phi,\varthet,\varph)(r^2 + \varr), \nonumber\\  
\sigma_{yz}^2  &=& A_6f_6(\varthet,\varph)g_6(\theta,\phi)r^2 + B_6h_6(\theta,\phi,\varthet,\varph)(r^2 + \varr),  
\end{eqnarray}
\end{footnotesize}where $A_1 \cdots A_6$ and $B_1 \cdots B_6$ are numerical constants, $f_1() .. f_6()$ are exponential functions of $\varthet,\varph$, $g_1() .. g_6()$ are trigonometric functions of $\theta,\phi$, and $h_1() .. h_6()$ are functions containing trigonometric and exponential expressions involving $\theta,\phi$ and $\varthet,\varph$, respectively. Exact functions can be obtained from equations (10) of\cite{l_mo_bar_shalom_unbiased_converted} or pp.\ $414$, $417$ of \cite{miller_coordinate_transform}.

\subsubsection{Assumptions in Our Scenario}

Restoring the superscripts and subscripts for the spherical and Cartesian coordinate person measurements, we make the following assumptions on the MAV observation model.

\textit{Bearing Measurement:} We employ a separate yaw controller \cite{eprice} to guide the yaw angle of MAVs towards the tracked person. This is to ensure that the MAV's camera is always oriented towards the tracked person. Due to this, we can assume that the measurement component $\phi_t^{P,k}$ is almost always approximately zero $(\phi_t^{P,k} \approx 0)$.

\textit{Inclination Measurement:} We assume that the person's inclination measurement $\theta_t^{P,k}$ remains close to $\pi/2$ $(\theta_t^{P,k} \approx \pi/2)$ as observed in  Fig.~\ref{fig:spher2cart}. 
The rationale for this assumption is as follows. Due to our formation controller (as described in subsequent sub-sections) the MAVs track and follow the person while maintaining a desired altitude and distance to her/him. Consequently, the change in $\theta_t^{P,k}$ is negligible. Hence, we assume $\theta_t^{P,k} \approx \pi/2$.

Substituting these approximate values for $\phi_t^{P,k}$, $\theta_t^{P,k}$ and $C_1(r_t^{P,k})^2$ for $\varr$ (see beginning of this section for the latter substitution) in (\ref{eqn:single_mav_model}) and using expressions from either \cite{l_mo_bar_shalom_unbiased_converted} or \cite{miller_coordinate_transform}, the elements of $\boldsymbol{\Sigma}_t^{P,k}$ are obtained in a compact form as,

\begin{footnotesize}
\begin{equation} \label{eq:covmodel}
\sigma_x^2 = C_x (r_t^{P,k})^2, ~\sigma_y^2 = C_y (r_t^{P,k})^2, ~\sigma_z^2 = C_z (r_t^{P,k})^2, ~\sigma_{xy}^2 = \sigma_{xz}^2 = \sigma_{yz}^2 = 0,
\end{equation}
\end{footnotesize}where $C_x,C_y,C_z$ are constants involving exponential functions of constants $C_1,C_2$ and $C_3$ only.

\subsubsection{Minimizing Fused Uncertainty for MAV Formations} 
\label{ideal_formation}

The minimum uncertainty in the person's fused state estimate is achieved when the following conditions are met.
	\begin{figure}[t]{}       
	\centering 
	\includegraphics[scale=0.38]{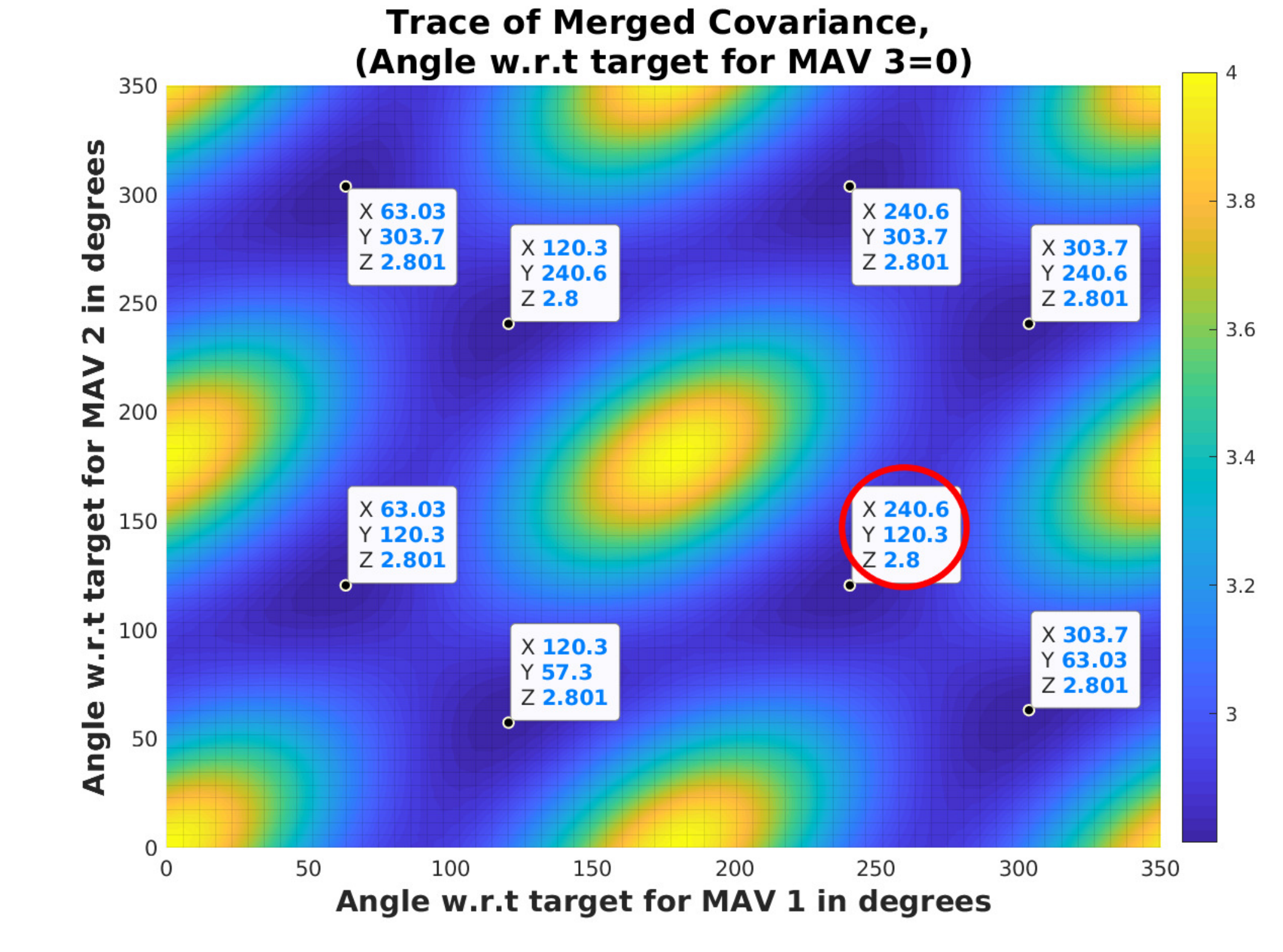}   
	\caption{Variation of the trace of fused covariance for different angular configurations.}     
	\label{f:merged_cov}
	\end{figure}	
\textit{(a) The angles between the MAVs about the person's estimated position are $\frac{2\pi}{K}$:}
Given measurements from observations of different robots (or sensors), \cite{bishop2010optimality} analytically derives optimal sensor geometries for unbiased and efficient passive target localization. Geometries that maximize the determinant of Fisher information matrix are identified. For K($\geq3$) independent measurements, the angle between sensors resulting in minimum fused uncertainty is either $\frac{2\pi}{K}$ or $\frac{\pi}{K}$. We validated this analysis by computing $\boldsymbol{\Sigma}_t^P$ from $\boldsymbol{\Sigma}_t^{P,k}$ for all possible angular configurations and using the recursive covariance merging equations from \cite{smith1986representation}. 
Fig.~\ref{f:act_angle} illustrates some of the possible angular configurations. Out of these, Fig.~\ref{f:act_angle} (a), (b) are examples of configurations that do not minimize $\boldsymbol{\Sigma}_t^P$. 
The angular configurations which minimize trace of $\boldsymbol{\Sigma}_t^{P}$ are highlighted in yellow boxes of Fig.~\ref{f:merged_cov}. 
Fig.~\ref{f:act_angle} (c) is one of those configurations which minimizes $\boldsymbol{\Sigma}_t^P$ and which we use in this work.  It is preferred over other optimal configurations because others do not ensure a uniformly distributed visual coverage of the tracked subject. This is important as our system is motivated by outdoor motion capture of an asymmetric subject (at least in one axis).

\textit{(b) The measurement uncertainty for each individual MAV is minimized:} The previous angular constraint condition minimizes the fused uncertainty for any given set of measurements from the MAVs. However, the fused covariance remains a function of each individual MAV's measurement uncertainty, $\boldsymbol{\Sigma}^{P,k}_t$ $\forall k\in[1\cdots K]$. $\boldsymbol{\Sigma}^{P,k}_t$ is a function of the relative position of MAV $k$ w.r.t.\ the tracked person. Notice that each MAV's position remains controllable without changing the angular configuration of the whole formation. Therefore, the fused uncertainty is completely minimized by minimizing the trace of $\boldsymbol{\Sigma}^{P,k}_t$ for each MAV $k$. How this is done is discussed in the following subsection.

\subsection{Decentralized Quadratic Model Predictive Control} \label{DQMPC_sec}
The trace of $\boldsymbol{\Sigma}_t^{P,k}$ obtained using (\ref{eq:covmodel}) is as shown below. 

\begin{footnotesize}
	\begin{equation}\label{tracecov}		
	tr(\boldsymbol{\Sigma}_t^{P,k})=(C_1  + C_2 + C_3)(r_t^{P,k})^2= \kappa (r_t^{P,k})^2,
	\end{equation}
\end{footnotesize}where $\kappa = C_1 + C_2 + C_3$. Minimizing (\ref{tracecov}) ensures that each MAV improves its measurement uncertainty. However, $r_t^{P,k}\approx0$ would cause the MAVs to collide with the tracked person and other MAVs. In order to ensure the safety of the person, we limit the MAV to reach a desired circular safety surface centered at the person with a radius $d_{des}$ at a desired altitude $h_{des}$ from her/him. Hence, instead of minimizing (\ref{tracecov}), which is equivalent to minimizing $(\x_t^{k}-\x_t^{P,k})^T \kappa\mathbf{I}_3 (\x_t^{k}-\x_t^{P,k})$, we now minimize $(\x_t^{k}-\xproj_t^{k})^T \kappa\mathbf{I}_3 (\x_t^{k}-\xproj_t^{k})$. The latter expression is in terms of the MAV's position $(\x_t^{k})$ and a desired position $(\xproj_t^{k})$ chosen on the aforementioned safety surface and lying in the direction from the MAV $k$ to the tracked person. Here, $\mathbf{I}_3$ is a $3\times 3$ identity matrix.
We now define the objective of our MPC running on each MAV k as follows. Minimize,
\begin{itemize}
	\item the distance to the desired target surface (safety surface) around the tracked person,
	\item the difference in velocity between the MAV and the fused velocity estimate of the tracked person and,
	\item MAV control effort.
\end{itemize} 
Consequently, the optimization objective of our active tracking MPC is as given below.

{\footnotesize \begin{equation}\label{cost}
	J_{\mathrm{ACT}} =  \sum_{n=0}^{N} (\bm{u}_t^{k}(n)^{\top}\bm{W}_E~\bm{u}_t^{k}(n)) \ + {\X^{k}_t}^{\top}\begin{bmatrix}
	\kappa\mathbf{I}_3 & 0 \\ 0 & W_{\dot{X}}
	\end{bmatrix}\X^{k}_t  
	\end{equation}
}where {\small $\X^{k}_t = [\x_t^{k}(N+1)^{\top} \ \dot{\x}_t^{k}(N+1)^{\top}]^{\top} -  [(\xproj_t^{k})^{\top} \ ({\dot{{\x}}}_t^P)^{\top}]^{\top}$}. The first part of this objective minimizes the input control signal and the second part ensures that the distance between desired terminal state (position and velocity) {\small$[\xproj_t^{k} \  {\dot{{\x}}}_t^P]$} and the final horizon robot state {\small$[\x_t^{k}(N+1) \ \dot{\x}_t^{k}(N+1)]$} is minimized. 
In order to  ensure continuous and smooth person tracking, we minimize the difference in velocity of the MAV $(\dot{\x}_t^{k})$ and the fused velocity estimate of the person $({\dot{\x}}_t^P)$ available from the EKF. 
$\kappa$ is as defined in (\ref{tracecov}) and is experimentally determined. $\mathbf{W}_E\in\mathbb{R}^{3\times3}$ and $\mathbf{W}_{\dot{X}}\in\mathbb{R}^{3\times3}$ are custom-defined diagonal positive semi-definite weight matrices.  
Furthermore, the active tracking MPC is subjected to the following constraints.
\begin{itemize}
\item    A non-convex constraint of maintaining a desired angular difference of $\frac{2\pi}{K}$ w.r.t.\ other MAVs about the fused estimate of the person's position ${\x}_t^P$.
\item   A non-convex constraint to maintain at least a distance $d_{min}$ from other obstacles.
\item     Regional and control saturation bounds for MAV states and control inputs for a horizon.
\end{itemize}
The non-convex constraints are embedded into the MPC dynamics as repulsive potential field forces $\mathbf{f}_t^{k}$ \cite{rahult}. The computation of these forces is discussed in Sec. \ref{non_convex}.
The convex MPC is thus defined by the following equations.

{\footnotesize
\begin{equation} \label{DQMPC}
	\x_t(1)^{k*}\dots\x_t(N+1)^{k*},\ao_t^{k*}(0)\dots\ao_t^{k*}(N) = \argmin{\ao_t^{k}(0)\dots\ao_t^{k}(N)} (J_{ACT})
\end{equation} \vspace{-1em} \begin{align}
	 \text{s.t} \quad & \left[\x_t^{k}(n+1)^{\top} ~ \dot{\x}_t^{k}(n+1)^{\top}\right]^\top = \nonumber\\
	& \quad\quad\quad\mathbf{A}\left[\x_t^{k}(n)^{\top} ~ \dot{\x}_t^{k}(n)^{\top}\right]^\top + \mathbf{B}(\ao_t^{k}(n)+\mathbf{f}_{t}^{k}(n)+\mathbf{g}),  \label{state-space} \\ 
	& [\x_\mathrm{min}^T ~ \dot{\x}_\mathrm{min}^T] \leq ~ [\x_t^{k}(n)^T ~ \dot{\x}_t^{k}(n)^T] \leq ~ [\x_\mathrm{max}^T ~ \dot{\x}_\mathrm{max}^T], \\
	& \hspace{1.6cm} \ao_\mathrm{min} \leq ~ \ao_t^{k}(n) \leq \ao_\mathrm{max}
\end{align}}
The 3D translation motion of MAV $k$ is governed partly using accelerations $\ao_t^{k}(n) = \ddot{\bm{x}}_t^{k}(n)$ and partly by an external control input $\mathbf{f}_t^{k}(n)$, where $n$ is the horizon time step. The state-space translational dynamics of the robot is given by \eqref{state-space}. Dynamics ($\mathbf{A} \in \mathbb{R}^{6\times6}$) and control transfer ($\mathbf{B} \in \mathbb{R}^{6\times3}$) matrices are given by

\begin{footnotesize}
	\begin{equation}
	\mathbf{A}=
	\begin{bmatrix}
	\mathbf{I}_3 & \Delta t\mathbf{I}_3 \\
	\mathbf{0}_3 & \mathbf{I}_3
	\end{bmatrix}, ~
	\mathbf{B}=
	\begin{bmatrix}
	\frac{\Delta t^2}{2}\mathbf{I}_3 \\
	\Delta t\mathbf{I}_3
	\end{bmatrix},
	\end{equation}
\end{footnotesize}where $\Delta t$ is the sampling time.  $\mathbf{f}_t^{k}(n)$ is a real-time computed external control input representing non-convex constraints of formation configuration and obstacle avoidance. The next section details the computation of these forces.

Note that in our formulation optimality is regarding the minimization of uncertainty in the person's 3D position jointly estimated by all MAVs. The way we formulate our MPC it cannot optimize the trajectories over the full belief space and hence transient behavior is not explicitly addressed. In our case, however, the MPC prediction time horizons are small enough (15 time steps or 1.5 seconds in total) to neglect the transient effects, as long as the MAVs are not very far away from the tracked person. Moreover, in the MPC we also minimize the difference between target person velocity and the MAV velocities, which further facilitates convergence to the desired view points.

Also, the optimal configuration may not always be feasible, especially, when the obstacles completely block or occlude the desired safety surface around the tracked person. This is inherent to our formulation because we enforce that obstacle avoidance and feasibility of trajectories take precedence (by considering them as hard constraints in MPC) over the optimality of the view points. This is important for the safety of aerial systems.

\subsection{Computation of External Control Input ($\mathbf{f}_t^{k}(n)$)} \label{non_convex}

Inter-MAV angular configuration constraints involve inverse-tangent operations with positive or negative arguments, making the constraint non-convex.
Moreover, avoiding collisions by maintaining a minimum distance w.r.t.\ other MAVs and obstacles in the environment, constraint the robot to operate in non-convex regions. In our previous work \cite{rahult}, we introduced the concept of converting non-convex constraints into external control inputs $\mathbf{f}_t^{k}(n)$ using repulsive potential fields.   
In this work, we use it to enforce formation angular configuration and collision avoidance constraints. The repulsive potential field function used in this work is referred to as cotangential field function (see (17) in \cite{tallamraju2019active}). 
The magnitude of the cotangential field force is a hyperbolic function denoted as $F^{k,j}(d)$. Argument $d$ is either the Euclidean distance or the absolute angular difference between MAV $k$ and entity $j$ (another MAV or obstacle). 
Let $d_{max}$ and $d_{min}$ be threshold radii, where $d_{max}$ defines the region of influence of the potential field. $d_{min}$ is the distance below which the potential field value tends to infinity. Practically, this is clamped to a large positive magnitude $F_{max}$ which is derived to guarantee obstacle avoidance (see Sec. \ref{sec:obs_guarantee}).

\subsubsection{Active Tracking Input}

To satisfy the inter-MAV angular configuration constraints, each MAV $k$ computes its angle about the tracked person in the world frame $(\gamma_t^{k}(n))$ and the angle of neighboring MAV $j$ about the tracked person $(\gamma_t^{j}(n))$. 
The absolute angular difference defined as $d_{act}(n) =|\gamma_t^{k}(n) - \gamma_t^{j}(n)|$, is used as the argument to compute the potential field magnitude.  
We compute for MAV $k$ and teammate MAV $j$,

{\footnotesize \begin{equation} \label{act_force_mag}
F_{act}^{k,j}(n) = (|r_{des} - \|\x_t^{k}(n)-{\x}_{t}^{P}\|_2| + c)  \;F^{k,j}(d_{act}(n)) \;\;\forall j
\end{equation}}
The factor {\small$|r_{des} - \|\x_t^{k}(n)-{\x}_{t}^{P}\|_2|$} in (\ref{act_force_mag}) helps avoid field local minima. This is because, if two robots have similar angles of approach (i.e., small $d_{act}(n)$), the MAV farther away from the desired distance $(r_{des})$ is repelled with a higher force than the MAV near the desired distance (see \cite{rahult} for a detailed explanation on how these forces avoid field local-minima problems associated with potential field based planners). Factor $c$ is a small positive constant which ensures that the force magnitude is non-zero at the target surface if the desired angular difference is not yet achieved. $F_{act}^{k,j}(n)$ acts in a direction $\mathbf{\alpha}^{k,j}_\perp$ which is normal to the direction of approach to the tracked person $\mathbf{\alpha}^{k,j} =  \frac{\x_t^{k}(n)-\x_{t}^{P}}{\|\x_t^{k}(n)-\x_{t}^{P}\|_2}$. In the plane of approach towards the person, there are two direction choices for a MAV, namely, $\pm \mathbf{\alpha}^{k,j}_\perp$. The direction pointing away from neighboring MAV $j$'s position is chosen. This choice ensures a natural deadlock resolution for robots having similar angles of approach to the tracked person.  The total active tracking external control input of MAV $k$ w.r.t. all other teammate MAVs is 

{\footnotesize
\begin{equation}\label{actforce}
\mathbf{f}_{act}^{k}(n) = 
\sum_{j}{F_{act}^{k,j}(n)\alpha^{k,j}_\perp}
\end{equation}}
\normalsize 
For the active tracking control input $d_{max} = (\frac{2\pi}{K})$.

\smallskip

\subsubsection{Obstacle Avoidance Input}

\textit{\\Dynamic Obstacle Avoidance}:
All the teammates of MAV $k$ are considered as dynamic obstacles. The argument $d$ for $F^{k,j}(d)$ in case of dynamic obstacle avoidance is the euclidean distance between MAV $k$ and teammate MAV $j$. It is defined as $d_{dyn}(n) = \|\x_t^k(n) - \x_t^j(n) \|_2, \forall n > 0$. Magnitude of $F^{k,j}(d)$ is computed using MAV $k$ and teammate $j$'s horizon motion plans. This is enforced along the direction $\mathbf{\beta}^{k,j} = \frac{\x_{t}^{k}(n)-\x_t^{j}(n)}{\|\x_{t}^{k}(n)-\x_t^{j}(n)\|_2}$, which is a unit vector pointing in the direction away from the teammate MAV $j$'s horizon motion plan.
The total dynamic obstacle avoidance external input  $\mathbf{f}_{dyn}^{k}(n)$ is as follows.

\begin{footnotesize}
\begin{equation}\label{rep_dyn}
	\mathbf{f}_{dyn}^{k}(n) = 
		\sum_jF^{k,j}(d_{dyn}(n))\;\mathbf{\beta}^{k,j}
\end{equation}
\end{footnotesize}
The choice for $d_{max}$ and $d_{min}$ for dynamic obstacle avoidance are discussed in the following sub-section \ref{sec:obs_guarantee}.

\textit{Static Obstacle Avoidance}:
The total external control input due to the $M$ static obstacles is as given below.

\begin{footnotesize}
\begin{equation}\label{rep_sta}
	\mathbf{f}_{sta}^{k}(n) = 
	\sum_jF^{k,m}(d_{sta}(n))\;\mathbf{\beta}^{k,m},
\end{equation}
\end{footnotesize}where $d_{sta}(n) =  \|\x_t^k(n) - \x_t^m \|_2$.
The tracked person is considered as an additional static obstacle for each MAV. An external control input $\mathbf{f}^k_{target}(n)$ is computed so as to enforce her/his safety. 
Further, multiple static obstacles placed closed together might cause the MAV $k$ to get stuck in a field local minima (as shown in \cite{rahult}). To avoid these scenarios, we compute a force $\mathbf{f}_{ang}^{k}(n)$ to penalize MAV approach angles ($\gamma_t^k(n)$) which have static obstacles in the line of approach. The computation of this force is similar to the active tracking force described in the previous section. 
The total external control input for obstacle avoidance $\mathbf{f}_{obs}^{k}(n)$ is given by the summation $\mathbf{f}_{obs}^{k}(n) = \mathbf{f}_{dyn}^{k}(n) + \mathbf{f}_{sta}^{k}(n) + \mathbf{f}_{target}^{k}(n) + \mathbf{f}_{ang}^{k}(n)  , \forall n$

Subsequently, the total external control input $\mathbf{f}_{t}^{k}(n)$ acting on MAV $k$ is determined as follows.

{\footnotesize \begin{equation} \label{clamping}
	\mathbf{f}_{t}^{k}(n)=
	\begin{cases}
	\mathbf{f}_{act}^{k}(n) + \mathbf{f}_{obs}^{k}(n)  & \text{if} ~~ \|\mathbf{f}_{act}^{k}(n) + \mathbf{f}_{obs}^{k}(n)\|_2 < F_{max} \\
	F_{max}\frac{\mathbf{f}_{act}^{k}(n) + \mathbf{f}_{obs}^{k}(n) }{\|\mathbf{f}_{act}^{k}(n) + \mathbf{f}_{obs}^{k}(n)\|_2} & \text{if} ~~  \|\mathbf{f}_{act}^{k}(n) + \mathbf{f}_{obs}^{k}(n)\|_2 \geq F_{max}         
	\end{cases}
	\end{equation}}

Next, we describe the considerations for choosing $d_{min}$, $d_{max}$ and $F_{max}$ to guarantee obstacle avoidance.

\subsubsection{Obstacle Avoidance Guarantee} 
\label{sec:obs_guarantee}
To ensure stable external control inputs and guarantee obstacle avoidance, we take into account the following important considerations.
\begin{itemize}
	\item maximum tracking error ($e_{max}$) of low-level controller w.r.t generated way-point $(\x_t^{k})$ for one time step $\Delta t$,
	\item maximum magnitude of the MAV velocity $(\|\dot{\x}_\mathrm{max}\|_2)$,
	\item self-localization uncertainty of the MAV $(\boldsymbol{\Sigma}_t^k)$,
	\item localization uncertainty of teammate MAVs $(\boldsymbol{\Sigma}_t^j)$,
	\item communication or packet loss with teammate MAVs.
	\item maximum cotangential force magnitude $F_{max}\geq \|\ao_{max}\|_2$.
\end{itemize} 
$d_{min}$ and $d_{max}$ corresponding to the external control input for obstacle avoidance are defined as follows. 
\begin{eqnarray} \label{guarantee}
	d_{min} = e_{max} + \|\dot{\x}_{max}\|_2\Delta t, ~	d_{max} = r_{\sigma_t}^{k} + r_{\sigma}^{j} + r_{comm,t}^{k, j} 
\end{eqnarray}
where, $r_{\sigma_t}^{k}$ is the maximum eigenvalue of $\boldsymbol{\Sigma}_t^k$. $r_{\sigma}^{j}$ is the experimentally determined maximum possible eigenvalue of the localization uncertainty $(\boldsymbol{\Sigma}_t^j)$ of the teammate MAV $j$. Wireless communication delays are unavoidable in real-time implementations and cannot be ignored as they affect the MAVs knowledge of teammate MAV positions. $r_{comm,t}^{k, j}$ accounts for it by increasing in proportion with the communication delay between the MAVs. $r_{comm,t}^{k, j} \propto \|\dot{\x}_\mathrm{max}\|_2\Delta t$, which is the maximum distance a neighboring MAV can travel in $\Delta t$ seconds.

%% file: experiments.tex
\section{Experiments and Results} \label{sec:results}
\subsection{Real Robot Experiments}
\label{sec:exps}

\subsubsection{Setup}
We use three self-developed and customized octocopter MAVs in the field experiments. 
The MAVs are equipped with an Intel i7 CPU, an NVidia Jetson TX1 GPU and Open Pilot Revolution flight controller system on chip.
Each MAV runs the CDT algorithm (Fig.~\ref{CDT}) on-board and in real-time, including the deep neural network-based single shot detector (SSD-Multibox). 
The MAVs use on-board GPS and IMU data, acquired at 100Hz, for localization. We place Emlid Reach differential GPS receiver on each MAV and on either shoulder of the tracked person to acquire ground truth (GT) location. The data from the differential GPS is not used during real-time experiments and stored only for comparison. 
For a decentralized implementation, a ROS multi-master setup with communication over Wifi is used. Each MAV continuously captures and stores camera images acquired at $40Hz$. The MPC is evaluated at $100$ Hz using the CVXGEN convex optimization library. The horizon length is $N = 15$ time steps corresponding to $1.5$ seconds. The state and velocity limits of each MAV is $[-20, -20, 3] \leq \x_t^{k} (n) \leq [20, 20, 10]$ in $m$ and $[-5, -5, -0.5] \leq \dot{\x}_t^{k} (n) \leq [5, 5, 0.5]$ in $m/s$. The control limits are defined 
as $[-3, -3, 3] \leq \bm{u}_t^{k} (n) \leq [3, 3, 11]$.  The desired horizontal distance to the tracked person in all experiments is set as $d_{des} = 8m$ and the desired height as $h_{des} = 8m$.  The tracking error w.r.t.\ to the MPC generated way-point is $\epsilon_{max} \approx 1m$, therefore we use $d_{min} = 1.5$m (see subection~\ref{sec:obs_guarantee}). $d_{max}$ w.r.t.\ teammate MAVs varies with the change in self localization uncertainty and communication losses. The task for each MAV in all experiments is to track and follow the person using the proposed approach. 
We conducted $5$ real-MAV experiments (\textbf{RE}), each with a $3$-MAV formation as follows. 

\begin{itemize}
	\item \textbf{RE~1}: Approach of Price et al. \cite{eprice} (duration 257s)
	\item \textbf{RE~2}: DQMPC (Tallamraju et al.) \cite{rahult} (231s)
	\item \textbf{RE~3}: Our approach (269s)
	\item \textbf{\mbox{RE 4}}: Our approach with one MAV avoiding emulated virtual obstacles (278s)
	\item \textbf{\mbox{RE 5}}: Our approach with all MAVs avoiding emulated virtual obstacles (106s)
\end{itemize}

\begin{figure}[!htp]
\centering
	\includegraphics[width=1.0\columnwidth]{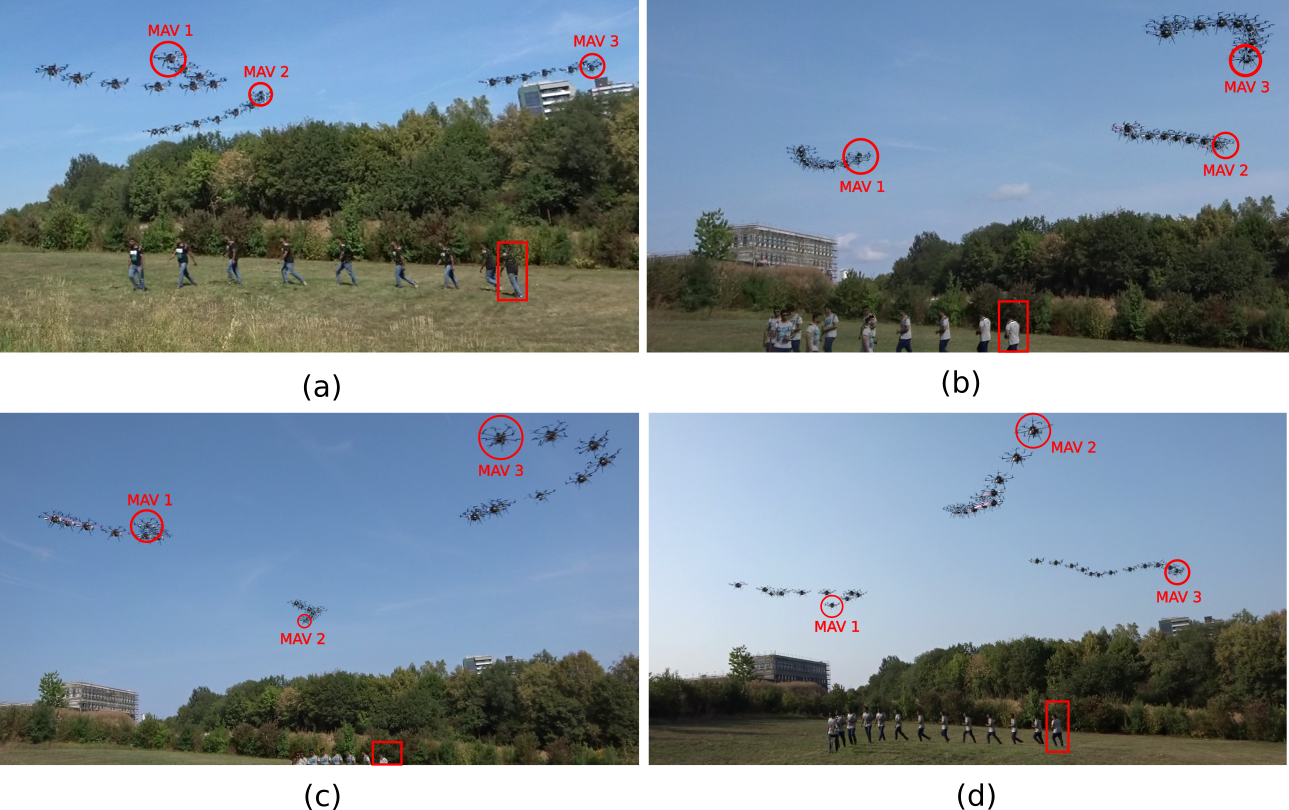}
	\caption{Multi-exposure images of short sequences from (\textbf{RE~1} to \textbf{RE~4}). (a) showcases the results based on our previous work \cite{eprice}. Notice that the MAVs are close to the target person and never uniformly spread around the person's position. (b) shows the results based on our previous work \cite{rahult}. Notice that the MAVs $2$ and $3$ are quite close to each other and the resulting formation is non-optimal for uncertainty minimization. In (c), (d) and the image in Fig.~\ref{cover} the results of the approach presented in the current paper is showcased. These correspond to experiments \textbf{RE~3} to \textbf{RE~5}, respectively. Notice that the MAVs are almost uniformly spread around the person's position and maintain an angular configuration with a difference of approx.\ $\frac{2\pi}{3}$ w.r.t.\ each other. Moreover, the MAVs successfully maintain the desired safe distance and altitude from the person. This configuration ensured minimization of fused uncertainty of the person's position in \textbf{RE~3}--\textbf{RE~5}. The experimental footage is provided in the accompanying video.}
\label{fig:snapshots}
\end{figure}

\begin{figure*} \centering
	\begin{tabular}{ cc }
	\includegraphics[width=1.3\columnwidth,  trim={0cm 0.1cm 0 0}, clip]{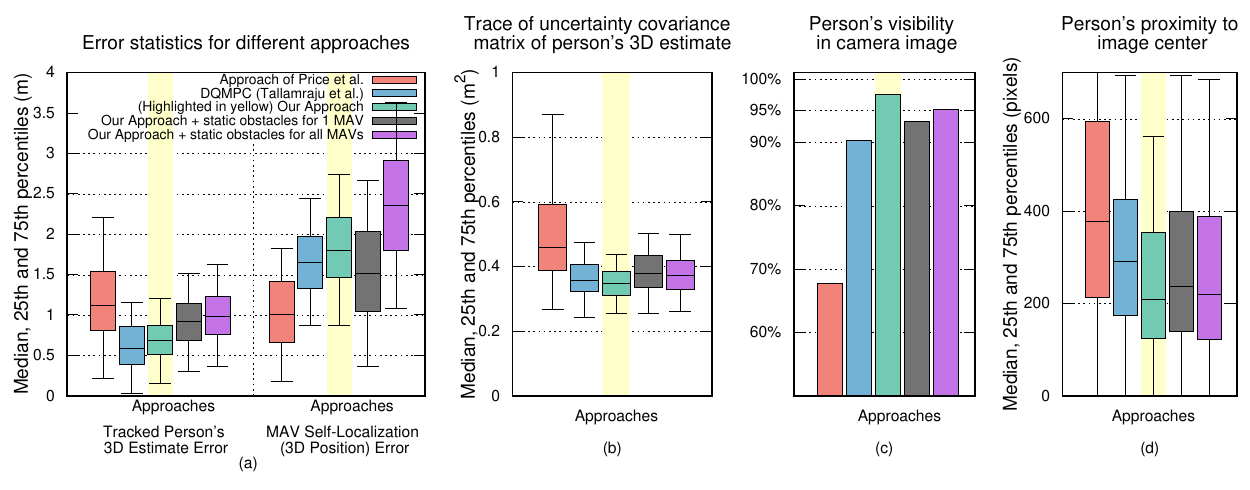} & \includegraphics[width=.67\columnwidth,  trim={0cm 0cm 0 0cm}, clip]{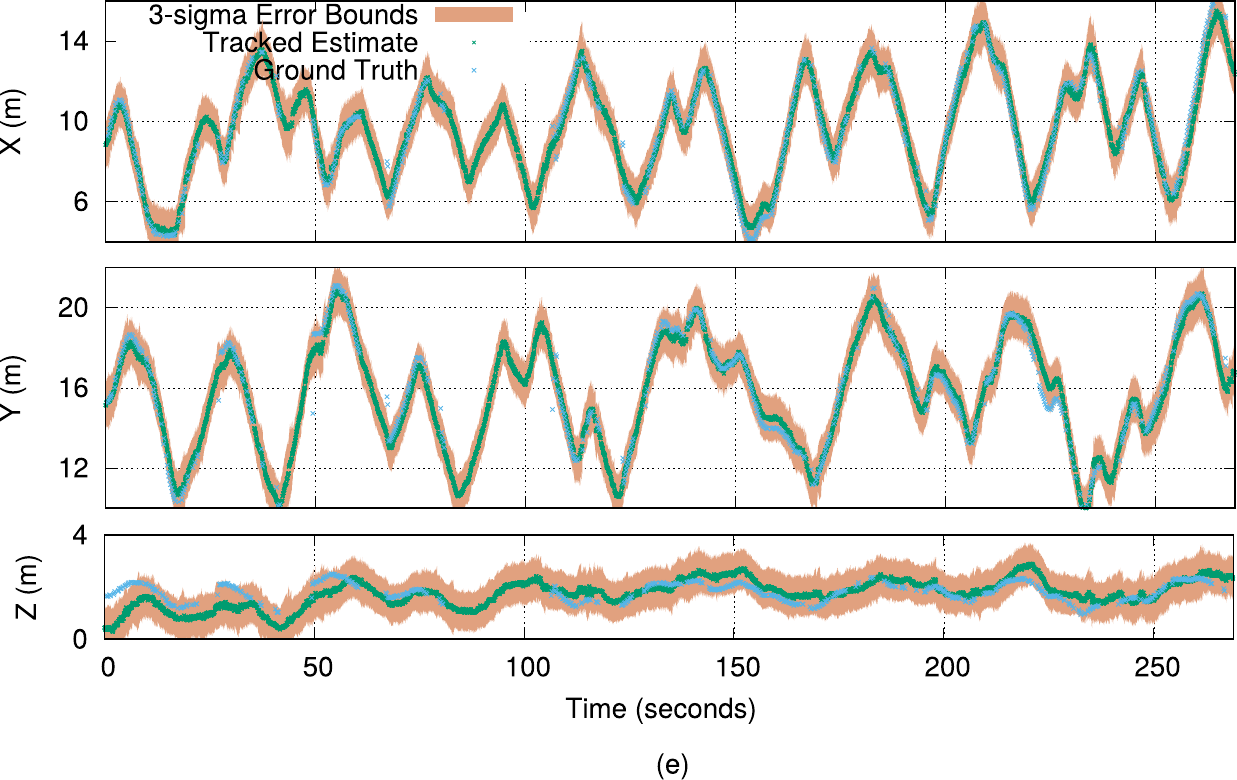}
\end{tabular}
	\caption{Comparison of different methods in real world experiments: (a) accuracy of the tracked person and MAV self-pose estimates; (b) uncertainty in the tracked person estimate; (c) fraction of total experiment duration (jointly for all MAVs) the tracked person was completely in the camera image frame; (d) distribution of distance of tracked person from image center. (e) Real world experiment \textbf{RE~3}: Tracked person's estimated (blue) and ground truth (green) trajectory comparison of our active perception approach (without simulated stationary obstacles). Ground truth remains within the 3$\sigma$ error bound of tracked estimate throughout the experiment duration except for a few initial seconds.}
\label{tracking}
\end{figure*}

In each experiment, the person remains stationary at first. Data collection starts after all MAV's have acquired the visual line of sight to the person and converged on a stable formation around him/her. The person then walks randomly at moderate speeds, runs or performs standard exercise movements (see attached video or here\footnote{\url{https://youtu.be/0Al3MlwOR1I}} for details). In the end, MAVs are manually landed. The error in the tracked position estimate of the person is calculated as the 3D Euclidean distance between the estimated and corresponding GT provided by the differential GPS system. Error for MAV pose is calculated similarly.

\subsubsection{Analysis of results}

Images in Fig.~\ref{fig:snapshots} and Fig.~\ref{cover} show short sequences from \textbf{RE~1} -- \textbf{RE~5}.
Fig.~\ref{tracking}(e) compares the corresponding GT with the tracked person's position estimate obtained by one of the MAVs using our proposed approach in \textbf{RE 3}.
We analyzed the estimate's accuracy, uncertainty, and the MAV self-pose accuracy for all experiments in Fig.~\ref{tracking}(a-b). {RE~3} achieves significantly more accurate person track estimates (mean error of $\sim 0.7$m) than using state of the art method in \textbf{RE~1} (mean error of $\sim 1.2$m), despite the worse self pose estimates of the MAVs in \textbf{RE~3} (the self-localization errors were due to the high errors in the MAV's GPS localization and may vary arbitrarily over different experiments). This showcases the ability of our approach to attain high tracking accuracy despite bad GPS reception. 

Even though the state of the art approach in \textbf{RE~2} achieves similar accuracy to that of our approach in \textbf{RE~3}, we see that our approach outperforms the former in keeping the person always in the camera image and close to the image center as shown by bar and box plots in \mbox{Fig.~\ref{tracking}(c-d)}. Moreover, our approach achieves least tracking uncertainty in the tracked person estimate as compared to the other two state-of-the-art methods.
The presence of obstacles that the MAVs need to navigate around affects the target tracking and self pose accuracy in \textbf{RE~4} and \textbf{RE~5}. This is mainly due to the additional maneuvering overhead to avoid obstacles. However, the ability to keep the person close to the center of the image is only slightly affected.
Fig~\ref{tracking}(c-d) compares the different approaches ability to keep the tracked person centered in each MAV's camera view. Our approach in \textbf{RE~3} not only reduces the average distance between the tracked person and each camera's image center but also ensures the person is completely covered by the camera image in $97.5\%$ of camera frames. This is a crucial feature for a system designed for aerial motion capture.

%% file: sim_experiments.tex
\subsection{Simulation Experiments}
\label{sec:sim_exps} 
\subsubsection{Setup}

The proposed algorithm was simulated in the Gazebo. The simulations were conducted on a standalone Intel i7-3970X CPU with an NVIDIA GTX 1080 GPU. We used AscTec Hexarotor Firefly MAVs in a world of $20m \times 20m \times 20m$. Each MAV has a rigidly attached Asus Xtion camera with their parameters set to the real MAV camera parameters. We simulate GPS and IMU drift by imposing random-walk offset on the ground truth position. We simulate a human actor model and guide it along a pre-defined trajectory. The actor traverses on a randomly varying terrain interspersed with emulated trees. The trees create obstacles and occlusions for the hovering MAVs.

\subsubsection{Comparison of methods}

\begin{figure*}[!htp]
\centering
\includegraphics[width=1.3\columnwidth,  trim={0cm 0.1cm 0 0}, clip]{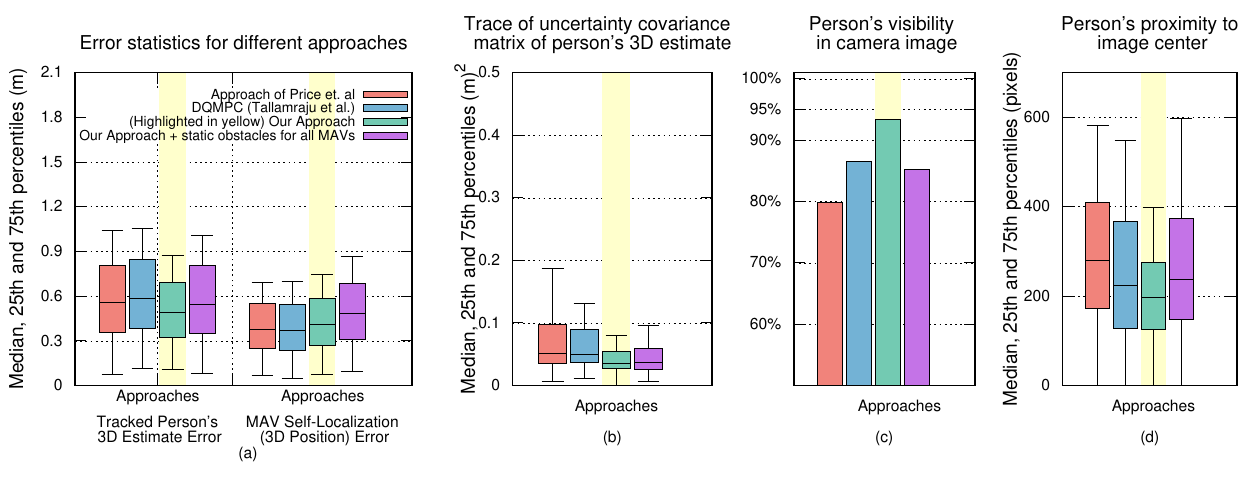} \hspace*{5pt} \includegraphics[width=.67\columnwidth,  trim={0cm 0cm 0 0cm}, clip]{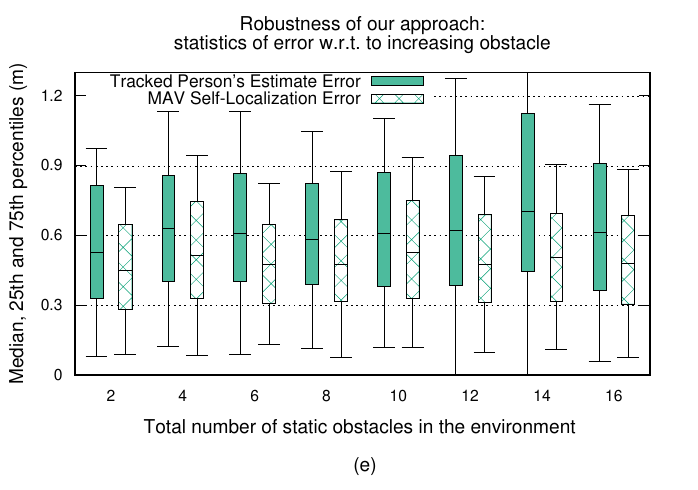}
\caption{Statistics of Simulation Experiment: (a)--(d) please refer caption of Fig.~\ref{tracking} for description of these plots; (e) robustness of our approach with increasing static obstacles.}
\label{comp_sim}
\end{figure*}

For each of the following methods, we conducted 30 simulation runs with a $5$ MAV formation: (i) \textbf{SE~1}: Approach of Price et al. \cite{eprice}, (ii) \textbf{SE~2}: DQMPC (Tallamraju et al.) \cite{rahult}, (iii) \textbf{SE~3}: Our approach, and (iv) \textbf{\mbox{SE 4}}. Our approach with all MAVs avoiding emulated virtual obstacles.
First 10 runs were of 120 seconds each, and next 20 runs were 180 seconds each. Figure~\ref{comp_sim} shows the statistics of this set of experiments. We clearly observe that our proposed approach with or without static obstacle avoidance outperforms both the state of the art approaches. Especially, when static obstacles are not present, our approach is able to keep the person fully in the camera image frame almost $94\%$ of the experiment duration. Also, it significantly outperforms all other methods in minimizing the joint uncertainty in the target position estimate.

\subsubsection{Scalability Experiments}

\begin{figure} 
	\centering
	\includegraphics[scale = 0.72]{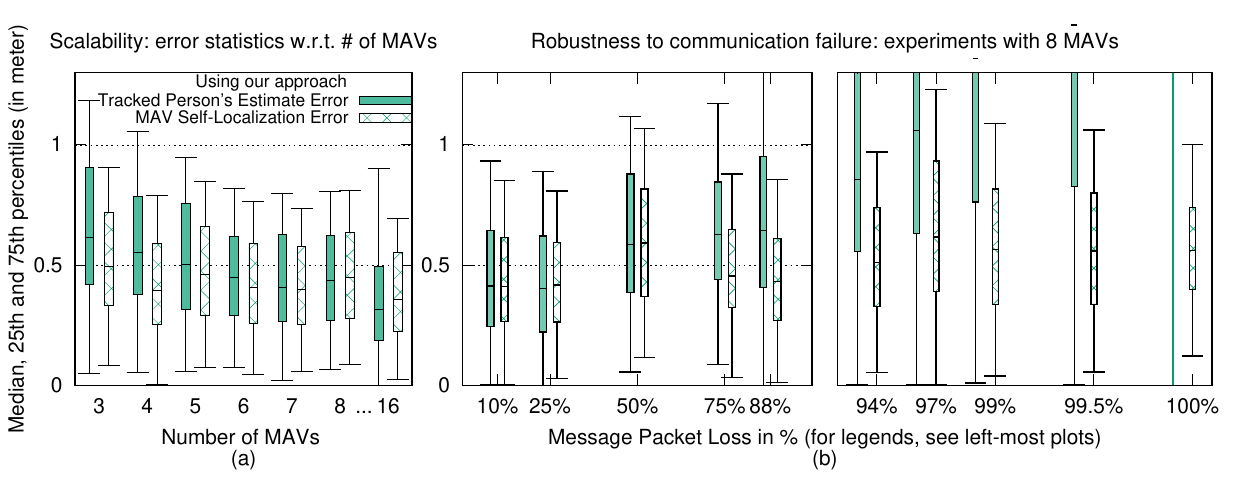}
	\caption{Our method's scalability and robustness.}
	\label{scale_network_sim}
\end{figure}

Fig.~\ref{scale_network_sim}(a) shows the result of target tracking with our method for an increasing number of robots. We conduct 10 simulation trials for each number of robot configuration. Real-time factor for all runs in this experiment was set to $0.1$. We validate the scalability of our model predictive controller by running these experiments in an environment with several static and known obstacles (in the form of trees) and perfect communication for a formation of 3 to 16 MAVs. We observe nearly linear improvement in tracking error with a higher number of robots. At the same time, we notice that computational requirements did not affect real-time performance for up to 16 robots.

\subsubsection{Experiments with Communication Failure}
In Fig.~\ref{scale_network_sim}(b), the effect of inter-robot communication failure on tracking is demonstrated for our approach through experiments with 8 MAVs and simulated static obstacles. Communication losses varying from $10\%$ to $100\%$ is simulated. The results were averaged over $3$ trials per communication loss percentage.  It can be observed that the tracking gets progressively worse with a higher percentage of communication loss. At $100\%$ communication loss, each robot relies only on its detection and does not cooperatively improve the target position estimate. Nevertheless, we observe that our approach is able to maintain an accurate target state estimate for up to 25\% communication loss.

\subsubsection{Experiments with increasing number of obstacles}
Fig.~\ref{comp_sim}(e) shows the results of our approach with increasing number of environmental obstacles. For each map, we conducted 5 trials (180 seconds each) with 5 MAVs tracking a randomly walking human model. We observe that by increasing the number of obstacles, the tracking error only slightly deteriorates. Moreover, the error in tracking is close to that of the environment with no static obstacles (see Fig. \ref{comp_sim}(a)). This result further validates our approach as the robots attain the desired configuration around the target person by navigating around randomly placed environmental obstacles.

%% file: conclusions.tex
\section{Conclusions and Future Work} \label{sec:conc}
We proposed a decentralized convex MPC-based algorithm for the MAVs to actively track and follow a moving person in outdoor environments and in the presence of static and dynamic obstacles.
In contrast to cooperatively tracking only the 3D positions of the person, the MAVs actively compute optimal local motion plans, resulting in optimal view-point configurations, which minimize the uncertainty in the tracked estimate. 
We showed how we embed all the non-convex constraints, including those for dynamic and static obstacle avoidance, as external control inputs in the MPC dynamics.
We evaluated our approach through rigorous and multiple real robot field experiments in an outdoor scenario. 
These experiments validate our approach and show that it significantly improves accuracy over the previous methods.
In simulations, we showed results with up to 16 MAVs. These demonstrate the scalability of our method as well as its robustness to communication failures. 
Future work includes addressing visibility issues, e.g., occlusions of the person and assymetry of the target, which are not explicitly solved by our current approach. We also intend to extend our method to full body human pose detection.